\documentclass{article} %
\usepackage{iclr2025_conference,times}

\iclrfinalcopy

\usepackage{amsmath,amsfonts,bm}

\def\eqref#1{equation~\ref{#1}}

\def\1{\bm{1}}

\def\vs{{\bm{s}}}

\DeclareMathAlphabet{\mathsfit}{\encodingdefault}{\sfdefault}{m}{sl}
\SetMathAlphabet{\mathsfit}{bold}{\encodingdefault}{\sfdefault}{bx}{n}

\usepackage{hyperref}
\usepackage{url}
\usepackage{booktabs}       %
\usepackage{amsfonts}       %
\usepackage{nicefrac}       %
\usepackage{microtype}      %
\usepackage{xcolor}         %
\usepackage{graphicx}
\usepackage{wrapfig,lipsum,booktabs}
\usepackage{multirow}
\usepackage{amssymb}
\usepackage{arydshln}
\usepackage{float}
\usepackage{soul}
\usepackage{color}

\usepackage{amsmath,amsfonts,bm}

\def\eqref#1{equation~\ref{#1}}

\def\1{\bm{1}}

\def\eg{\emph{e.g.}}
\def\ie{\emph{i.e.}}

\def\0{{\bf 0}}
\def\1{{\bf 1}}

\def\bV{{\bf V}}
\def\bW{{\bf W}}
\def\bX{{\bf X}}

\def\bm{{\bf m}}

\def\bs{{\bf s}}

\def\bx{{\bf x}}
\def\by{{\bf y}}

\def\vs{\emph{vs.}}

\newcommand{\tabincell}[2]{\begin{tabular}{@{}#1@{}}#2\end{tabular}}

\def\jing{\textcolor{black}}
\def\revise{\textcolor{black}}

\definecolor{citecolor}{HTML}{0071bc}
\definecolor{paleplum}{rgb}{0.8, 0.6, 0.8}
\hypersetup{
  colorlinks,
  citecolor=citecolor,
  linkcolor=red
}

\title{ME-Switch: A Memory-Efficient Expert Switching Framework for Large Language Models}

\author{
Jing Liu$^{1,2}\thanks{Work done during an internship at SenseTime Research.}$, Ruihao Gong$^{2,3}$, Mingyang Zhang$^{4}$, Yefei He$^{4}$,
Jianfei Cai$^1$, Bohan Zhuang$^1$\thanks{Corresponding author. Email: $\tt bohan.zhuang@gmail.com$} \\[0.2cm]
$^1$ZIP Lab, Monash University \quad  $^2$SenseTime Research \quad \\ $^3$Beihang University \quad
$^4$Zhejiang University \\
}

\begin{document}

\maketitle

\begin{abstract}
The typical process for LLM's development involves pre-training a general foundation model on massive data, followed by fine-tuning on task-specific data to obtain a series of specialized experts. Serving these experts can pose significant \revise{memory} challenges, as loading all experts onto devices is impractical, and frequent switching between experts in response to user requests can incur substantial I/O costs. 
Previous approaches decompose the expert weights as the pre-trained weights plus delta weights, followed by \jing{quantizing} the delta weights \jing{using output channel-wise step sizes} to reduce the model size. However, \jing{these methods overlook the fact that certain input channels of delta weights can cause significant quantization errors at extremely low bitwidths.}
Additionally, existing methods assume that the appropriate model for a user request is known in advance, which is not the case in practice. To this end, we introduce ME-Switch, a memory-efficient expert switching framework tailored for serving \revise{multiple LLMs}. To condense the number of bits required for describing the delta weights, we propose a salient-aware delta compression method that
\revise{first identifies which input channels of delta weights are salient based on reconstruction error and then}
employs mixed-precision quantization that selectively quantizes non-salient input channels of delta weights to extremely low bits while keeping the salient ones intact, significantly reducing storage demand while maintaining performance.
Moreover, we develop a model-level routing method that efficiently directs user queries to the most suitable expert by \revise{performing domain classification}. Extensive experiments show the promising memory efficiency and routing performance of ME-Switch. For example, when serving three models from the Mistral-7B family, ME-Switch reduces the model size by $1.74\times$ and maintains nearly lossless performance on instruction, mathematical reasoning, and code generation tasks. \jing{Furthermore, our method can efficiently serve 16 Mistral-7B models on \revise{an} NVIDIA A100 GPU.}
\end{abstract}

\section{Introduction}
Large language models (LLMs) such as GPT-4~\citep{achiam2023gpt} and Gemini~\citep{team2023gemini}, have achieved significant advancements in natural language processing (NLP). Trained on large-scale text datasets, these models develop a broad foundation of language understanding, enabling them to excel at tasks requiring common-sense knowledge. To acquire task-specific knowledge, LLMs can be further fine-tuned on specialized tasks, thereby adapting them for diverse applications such as interactive agents~\citep{touvron2023llama,jiang2023mistral}, code generation~\citep{luo2023wizardcoder, lozhkov2024starcoder}, and mathematical problem solving~\citep{luo2023wizardmath,lozhkov2024starcoder}, showing the remarkable versatility of LLMs. Nevertheless, given that no single model can master all tasks simultaneously, serving multiple LLMs, each precisely tailored for specific tasks, becomes essential for good performance.

\begin{figure}[t!]
	\centering
	\includegraphics[width=1.0\linewidth]{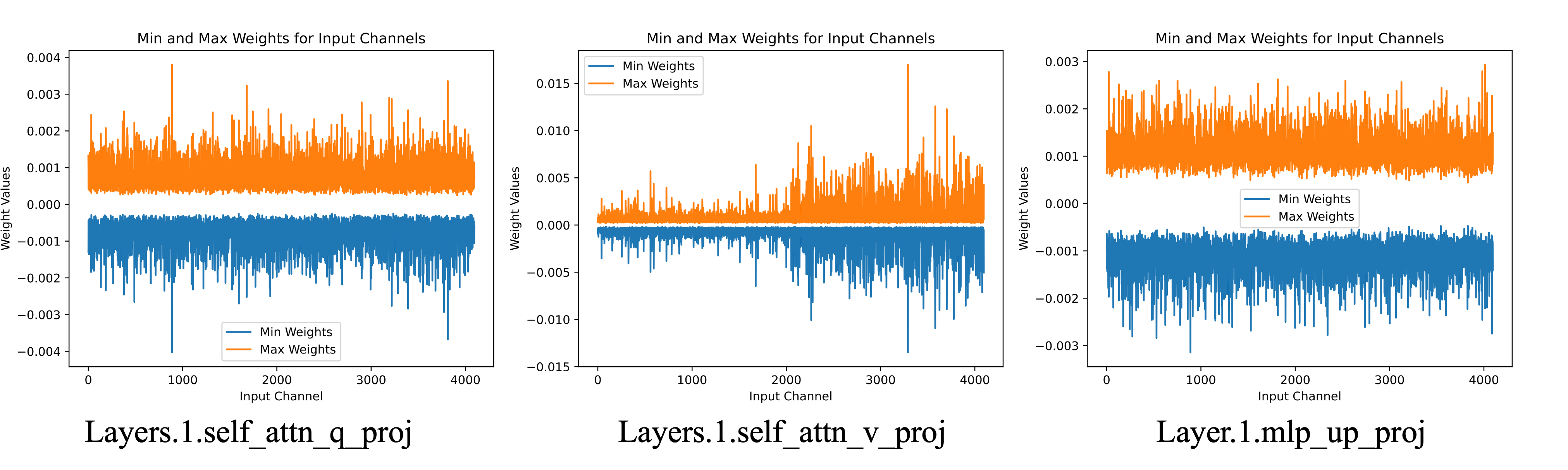}
        \vspace{-0.15in}
	\caption{An illustration of the input channel-wise maximum and minimum values for the delta weights of Speechless-Code-Mistral-7B. \jing{The variability across input channels highlights that certain salient channels, irrespective of their magnitude, can cause significant quantization errors when quantized with ultra low-bitwidth, which underscores their critical role in preserving performance.}}
	\label{fig:mistral_in_channel_min_max}
    \vspace{-0.25in}
\end{figure}

However, serving multiple models poses several major challenges. First, even with a relatively small number of models, the storage demands are significant due to the extensive number of parameters each model contains. For example, three LLaMA-2-70B models would collectively require over 384GB of storage, calculated as 128GB per model times three. Second, the substantial memory requirements of these models may make it impractical to load all of them into GPU memory simultaneously. While dynamically swapping model weights in and out of GPU memory as needed is feasible, the large size of the models makes this process slow and inefficient, significantly delaying response times and adversely affecting user experience.

To address the above challenges, \revise{existing methods~\citep{liu2024bitdelta,yao2023deltazip} decompose the weights of fine-tuned models into pre-trained weights and delta weights introduced during fine-tuning. While low-rank approximation \citep{hu2022lora} can compress these delta weights, it falls short for full fine-tuned models whose delta weights lack low-rank properties~\citep{liu2024bitdelta,lialin2024relora,hao2024flora}. As a result, prior work has adopted per-tensor quantization~\citep{liu2024bitdelta} as an alternative, significantly reducing storage needs and facilitating efficient sharing of the base model’s storage across multiple models.} 
\jing{Nevertheless, this method ignores the 
distinction
in delta weight values across different input and output channels, resulting in substantial information loss. To mitigate this issue, 
output channel-wise quantization~\citep{xiao2023smoothquant,wei2023outlier,liu2024qllm}
where each output channel is allocated its own learnable step size, still 
neglects input channel variations, as shown in Figure~\ref{fig:mistral_in_channel_min_max}.}
To further mitigate information loss, rescaling input channels before quantization can be employed, but this provides only limited alleviation at extremely low bitwidths.
\revise{In addition}, existing methods assume that the appropriate model to utilize in response to a user request is known in advance. In reality, \revise{dynamically selecting the optimal model is essential for effectively handling diverse queries.}
This gap highlights the need for an intelligent model selection mechanism that can identify the most suitable model based on specific user needs.

In this paper, we propose ME-Switch, a memory-efficient expert switching framework tailored for LLMs. \jing{To reduce the quantization error while simultaneously reducing storage needs,}
we develop a mixed-precision quantization method that quantizes non-salient input channels of delta weights to extremely low bits while \jing{preserving those salient ones, which could substantially increase quantization errors when quantized at very low bitwidths, in full precision.}
Since the number of salient input channels is relatively small, incorporating a limited amount of high-precision delta weights incurs negligible memory overhead and inference cost. To identify the important input channels of delta weights, one may select based on their magnitudes~\citep{dettmers2022gpt3}, as shown in Figure~\ref{fig:salient_weight_selection}(a), which however fails to capture those leading to high quantization error. In contrast, our approach identifies salient input channels of delta weights based on their impact on reconstruction errors in the output activations, as illustrated in Figure~\ref{fig:salient_weight_selection}(b). 
Additionally, we develop a routing method that selects the most suitable LLM in response to a user request, enabling efficient and accurate model switching. To simplify the routing problem, we assume that each LLM specializes in a distinct domain, such as code generation or mathematics reasoning. This allows us to treat model selection as a multiple-choice question-answering task where each option represents a target domain, effectively transforming the problem into a domain classification task, as shown in Figure~\ref{fig:model_routing}. Notably, dialogue LLMs exhibit strong instruction-following capabilities. Leveraging this, we propose using a small pre-trained dialogue LLM as a model-level router to solve the multiple-choice question-answering task. We then construct a multiple-choice question-answering dataset and fine-tune the router to improve its performance. Given its smaller size compared to the specialized LLMs, the router operates efficiently without adding significant overhead.

Our contributions can be summarized as follows. \revise{1) We introduce ME-Switch, a memory-efficient framework designed for serving multiple LLMs. It only stores a single full-precision pre-trained model and dynamically loads the appropriate compressed delta model weights in response to user queries.}
2) \revise{We develop a mixed-precision quantization method that significantly reduces the storage demands of serving multiple LLMs while maintaining performance, \revise{which is achieved by} selectively quantizing the non-salient input channels of delta weights and leaving the salient ones unchanged. We further introduce a routing method that dynamically selects the most appropriate LLM for a given query, which is accomplished by transforming the model selection problem into a domain classification task and then solving it with a small LLM.}
3) We conduct extensive experiments demonstrating the promising memory efficiency and routing performance of ME-Switch. \jing{Remarkably, when serving three models from the Mistral-7B family, ME-Switch not only delivers near-lossless performance on instruction, mathematical reasoning, and code generation tasks but also reduces the model size by $1.74\times$. More impressively, our method is able to serves up to 16 Mistral-7B models on a single NVIDIA A100 GPU without running out of memory.}

\section{Related Work}
\noindent\textbf{Efficient LLM serving.}
LLMs can be efficiently deployed on GPUs for high-throughput serving using several inference frameworks, such as vLLM \citep{kwon2023efficient} and Orca \citep{yu2022orca}. Given that a single LLM cannot excel across all domains, it is crucial to serve multiple LLMs simultaneously to handle diverse user queries effectively.
To determine which model to use during inference, Zooter \citep{lu2023routing} utilizes a task-level router to load different LLMs based on the incoming task requirements. 
This approach, however, introduces significant memory challenges, as hosting all LLMs on a GPU simultaneously can excessively strain the available VRAM. One solution is to serve multiple Low-Rank Adaptation (LoRA) modules within a multi-tenant serving system, such as Punica~\citep{chen2023punica} and S-LoRA~\citep{sheng2023s}. \revise{However, these methods fail to accommodate full fine-tuned models because such models typically do not display the low-rank characteristics necessary for LoRA approximation.}
Another viable strategy to address the above issue involves dynamically loading different LLMs from CPU memory to GPU memory as needed, thereby reducing peak GPU memory utilization. Nevertheless, this dynamic loading competes for GPU memory bandwidth with model-swapping operations, presenting a significant bottleneck. To mitigate these challenges, techniques like Deltazip \citep{yao2023deltazip} and BitDelta \citep{liu2024bitdelta} have been developed to compress the delta parameters. These methods allow task-specific delta parameters to be loaded into GPU memory on-demand, ensuring that only a single pre-trained model’s parameters reside permanently in GPU memory. This approach aims to achieve low-latency inference while minimizing the costs associated with maintaining numerous fine-tuned models in GPU memory. Our approach stands out by serving multiple LLMs with a minimal memory footprint with nearly lossless performance, which is achieved through an advanced delta weights quantization strategy and an adaptive model-level router.

\noindent\textbf{Delta compression.}
In recent years, numerous approaches have focused on reducing the storage overhead for maintaining different task-specific models through delta parameter compression. Model merging strategies often incorporate multiple tasks' delta parameters \citep{ding2023parameter} into the pretrained parameters to minimize the number of parameters needed for multi-task operations.
To address the parameter conflicts that often arise during model merging, Yu et al. \citep{yu2023language} proposed a method involving massive unstructured random pruning (achieving 90\% sparsity) of delta parameters and subsequent rescaling. This approach ensures that the accuracy of downstream tasks is not compromised. In a similar vein, Ties-merging \citep{yadav2024ties} developed a pruning strategy based on the magnitude and sign of delta parameters to further reduce conflicts and storage needs.
Additionally, the LoRA-based PEFT methods \citep{hu2021lora,valipour2022dylora,ping2024delta} introduce a novel approach by learning one or several low-rank matrices to represent the delta parameters. ZipLoRA \citep{shah2023ziplora} explores the sparsity within these low-rank matrices, allowing the fusion of low-rank matrices from different tasks to further reduce the spatial overhead of multi-task models. \revise{However, these methods struggle to accurately approximate delta parameters in full fine-tuned models, as these models usually lack low-rank properties~\citep{liu2024bitdelta,lialin2024relora,hao2024flora}.}
Conversely, some studies focus solely on compressing delta parameters for each task without merging, thereby mitigating performance degradation across multiple tasks. For example, Ryu et al. \citep{ryu2023efficient} combine quantization and low-rank estimation techniques to reduce the storage size of delta parameters.  Liu et al. \citep{liu2024bitdelta} push this further by quantizing each task's delta parameters to 1 bit, achieving more than a tenfold compression. However, these quantization methods often lack robust handling of outliers, resulting in performance declines compared to uncompressed delta parameters.
Moreover, models that do not integrate multi-task delta parameters require manual activation of specific delta parameters for each task, which reduces the model's applicability in multi-task environments.

\section{Preliminaries}

For simplicity, we employ uniform quantization~\citep{jacob2018quantization} to compress the models. Given a matrix $\bX$ with floating-point values (\eg, FP16), the quantization process can be expressed as:
\begin{equation}
    \label{eq:quantization}
    \begin{array}{ll}
            \hat{\bX} = \mathrm{quant}(\bX) = \mathrm{clamp} \left( \lfloor \frac{\bX}{s} \rceil, -Q_N, Q_P \right) \times s, 
    \end{array}
\end{equation}
where the function $\mathrm{clamp}(\bV, \bV_{\mathrm{min}}, \bV_{\mathrm{max}})$ clamps all elements in $\bV$ within the range $[ \bV_{\mathrm{min}}, \bV_{\mathrm{max}} ]$, the operator $\lfloor \cdot \rceil$ rounds a given value to the nearest integer, and $s$ is a learnable quantization step size initialized by $\max(| \bX |) / (2^{b-1}-1)$. Here, $Q_N$ and $Q_P$ denote the number of negative and positive quantization levels, respectively. For $b$-bit quantized weights, $Q_N$ and $Q_P$ are set to $2^{b-1}$ and $2^{b-1} - 1$, respectively. 
Since the rounding function is not differentiable, we use the straight-through estimator (STE)~\citep{Esser2020LEARNED} for gradient approximation following~\citep{Esser2020LEARNED}. For binary quantization, we use the quantization method following~\citep{rastegari2016xnor}.

Recent studies~\citep{liu2024bitdelta,yao2023deltazip} have shown that the weights of a fine-tuned model can be decomposed into the weights of the pre-trained model and the delta weights introduced during fine-tuning. Let $\bW \in \mathbb{R}^{m \times n}$ and $\bW_{\mathrm{FT}} \in \mathbb{R}^{m \times n}$ be the weight matrices of the pre-trained model and the fine-tuned model, respectively, where $m$ represents the number of input channels and $n$ denotes the output number of channels. The delta weight is defined as $\Delta = \bW_{\mathrm{FT}} - \bW$. To reduce storage requirements, one can perform quantization using Eq.~(\ref{eq:quantization}) to compress $\Delta$. 
However, this method often results in significant performance degradation because it assumes all delta weight channels are equally sensitive to quantization noise. 
Additionally, existing methods assume the appropriate model for a user request is predetermined. In practice, dynamically choosing the most suitable model based on the specific nature of each user's query is more effective.

\section{Proposed Method}
In this section, we propose ME-Switch, a memory-efficient expert switching framework designed to optimize the deployment of \revise{multiple} LLMs. \revise{ME-Switch reduces memory demands while maintaining performance by using mixed-precision quantization, which quantizes non-salient weights to extremely low bitwidths while keeping salient weights intact. Additionally,  ME-Switch introduces model-level routing to efficiently handle task-specific queries.} We will detail ME-Switch in the following sections.

\subsection{Salient-Aware Delta Compression}
\label{sec:salient_aware_delta_compression}
\jing{In this section, we introduce our salient-aware delta compression approach, which specifically targets the preservation of salient input channels during quantization. 
While quantization methods with learnable step sizes for each output channel can handle variations in output channels effectively~\citep{xiao2023smoothquant,liu2024qllm}, variations in input channels pose a significant challenge to maintaining model performance, as shown in Figure~\ref{fig:mistral_in_channel_min_max}. 
Some salient input channels of delta weights, regardless of their magnitude, can cause substantial information loss and degrade model performance when quantized to a very low bitwidth.} To address this, rescaling the input channels of delta weights before quantization~\citep{lin2023awq} offers a solution to mitigate some of the quantization error, \jing{which however provides only limited alleviation under the context of extremely low-bit quantization}. 

\begin{figure}[t!]
	\centering
	\includegraphics[width=0.97\linewidth]{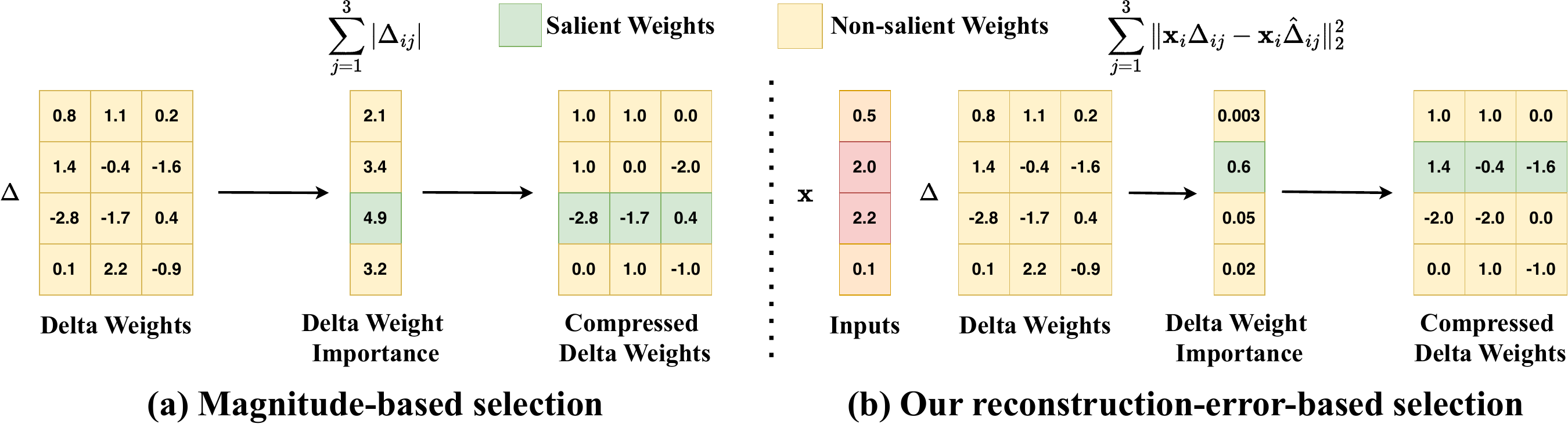}
        \vspace{-0.05in}
	\caption{An illustration comparison between the magnitude-based selection of salient delta weights and our reconstruction-error-based selection method. Given a delta weight matrix $\Delta \in \mathbb{R}^{m \times n}$, its quantized version $\hat{\Delta}$, and input $\bx \in \mathbb{R}^m$, where $m$ and $n$ denote the number of input and output channels, respectively, our method measures the importance of each input delta channel by $\sum_{j=1}^{n} \| \bx_i \Delta_{ij} - \bx_i \hat{\Delta}_{ij} \|_2^2$. %
 }
	\label{fig:salient_weight_selection}
    \vspace{-0.05in}
\end{figure}

\noindent\textbf{Salient-aware delta selection.} To protect salient input channels of delta weights, we develop a mixed-precision quantization method, where the majority of input channels of delta weights are quantized to low-bit precision, while only a small number of critical input channels \revise{are represented in their original precision.}
This approach achieves significant reductions in storage requirements with minimal performance loss. The remaining challenge lies in identifying these salient input channels. A naive approach might involve selecting the input channels with large magnitudes as the salient delta weights~\citep{dettmers2022gpt3}, as shown in Figure~\ref{fig:salient_weight_selection}(a). However, this method neglects the influence of input activations on the outputs, failing to identify channels that lead to high quantization error. To address this, inspired by the pruning metric~\citep{sun2024a}, we introduce a salient delta weights selection metric based on reconstruction errors in the outputs, considering both weights and input activations, as shown in Figure~\ref{fig:salient_weight_selection}(b). Specifically, given an input $\bx \in \mathbb{R}^{m}$, the reconstruction error for the input channel $i$ can be calculated by
\begin{equation}
    \begin{array}{ll}
    \sum_{j=1}^{n} \| \bx_i \Delta_{ij} - \bx_i \hat{\Delta}_{ij} \|_2^2,
    \end{array}
\end{equation}
where $\hat{\Delta}$ is the quantized delta weights and $n$ denotes the output channel number. With the reconstruction error defined, we choose those input channels with the top-$k$ largest reconstruction errors as the salient ones and retain them in 16-bit representation to maintain performance. 

\noindent\textbf{Efficient distillation.} The quantization step size in Eq.~(\ref{eq:quantization}) plays an important role in the final performance. To learn the quantization step size, we employ knowledge distillation to guide the alignment of the output logits of the quantized model with those of the full-precision fine-tuned model following~\citep{liu2024bitdelta}. To reduce training overhead, we freeze the delta weights and focus solely on optimizing the quantization step size $\bs$ using a small calibration dataset $\mathcal{X}$ by solving
\begin{equation}
\begin{array}{ll}
\arg\min_{\bs} \| f({\mathcal{X}}) - \hat{f}(\mathcal{X}; \bs) \|_2^2,
\end{array}
\end{equation}
where $f(\cdot)$ and $\hat{f}(\cdot)$ denote the output logits of the fine-tuned and quantized models, respectively. Thanks to the reduced number of trainable parameters, this training process is highly efficient.

\noindent\textbf{On-demand swapping.} The extremely low-bit compression for delta weights significantly reduces the model size, alleviating storage demands. This approach enables us to maintain a single pre-trained model while storing multiple sets of compressed delta weights, facilitating efficient on-demand swapping. In this scenario, the pre-trained model remains in GPU memory, and the corresponding compressed delta weights are loaded dynamically based on the user query. Compared with directly serving multiple models, our method is more GPU memory-efficient. Since LLM decoding is memory-bound~\citep{liu2023deja,sheng2023flexgen} due to the auto-regressive nature, reducing model size effectively decreases the parameter loading time, thereby improving decoding latency. To achieve fast inference, we decouple the matrix multiplication during inference into two components:
\begin{equation}
\begin{array}{ll}
    \label{eq:finetuned_inference}
    \by = \bx \bW_{\mathrm{FT}} = \bx (\bW + \Delta) \approx \bx \bW + \bx \Tilde{\Delta},
\end{array}
\end{equation}
where $\Tilde{\Delta}$ represents the compressed delta weights, including both the quantized unsalient and the full-precision salient delta weights. For $\bx \bW$, the computation is performed using a FP16 batched GEMM kernel. For $\bx \Tilde{\Delta}$, we implement an efficient Triton kernel~\citep{tillet2019triton} that fuses dequantization and matrix multiplication for efficient computation. \revise{The reduced memory footprint of the compressed delta weights enables our method to perform batched forward passes across multiple models simultaneously. This batching at the model level significantly enhances efficiency compared to the traditional approach, which processes each model individually—especially beneficial when serving multiple models (See Figure~\ref{fig:model_latency}).}

\subsection{Model-level Routing}
\label{sec:model_level_routing}
\begin{figure}[t!]
	\centering
	\includegraphics[width=0.95\linewidth]{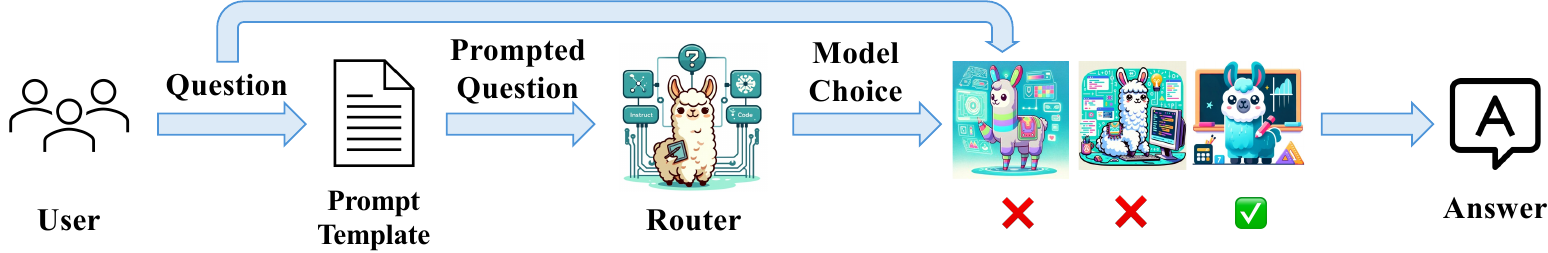}
        \vspace{-0.1in}
	\caption{An illustration of the model-level routing. \jing{We first prompt the model-level router with the user query using a template (See Section~\ref{sec:prompt_template} for more details) that presents a list of potential domains. The router then assesses these options and selects the most relevant domain by answering a multiple-choice question, effectively classifying the query into the corresponding category.}}
	\label{fig:model_routing}
    \vspace{-0.15in}
\end{figure}

\vspace{-0.5em}
Given a user query, 
we present a method to determine the appropriate model. Consider a set of LLMs represented as $\mathcal{F} = \{ f_1, f_2, \cdots, f_M \}$, where $M$ denotes the number of models. Given a user query $q$, we aim to find the most suitable LLM by solving the following problem $\arg\max_{f \in \mathcal{F}} P(q, f(q))$, where $P$ is a function that measures the quality or performance of the LLM response. To simplify the routing process, we assume that each LLM in $\mathcal{F}$ specializes in distinct domains such as code generation or mathematical problem solving.
This setup allows us to treat the routing challenge as a multiple-choice question-answering task, where each option corresponds to a specific domain, thereby transforming the problem into a domain classification problem. Note that dialogue LLMs like Qwen1.5-1.8B-Chat~\citep{qwen} exhibit capabilities in following instructions, which inspires us to utilize a small pre-trained LLM as a model-level router. As illustrated in Figure~\ref{fig:model_routing}, we first prompt the router with the user's question using a template designed to elicit domain classification. For the prompt template, please refer to Section~\ref{sec:prompt_template} of the appendix. Based on the router's response, we then dynamically load the corresponding compressed delta weights for the selected domain-specific model, such as a mathematical model, to generate outputs.

Since the router is not explicitly trained for query domain classification, its initial routing performance may be suboptimal. To improve the routing performance, we construct a multiple-choice question-answering dataset tailored for our routing problem. We collect instruction-following data from various domains and insert the query into the prompt template as shown in Figure~\ref{fig:model_routing}. The responses are constructed by considering the correct domain-specific model choice. We then fine-tune the router with our constructed dataset using supervised fine-tuning  to further improve the routing accuracy. 

\vspace{-0.5em}
\subsection{Discussions}
\label{sec:discussions}
\vspace{-0.5em}
\noindent\textbf{Comparisons with routing in Mixture of Experts (MoE).} Besides model-level routing, another approach to handle diverse user queries efficiently is to construct a MoE using a set of pre-trained expert models. This can be achieved by integrating the feedforward layers from all pre-trained LLMs into a single MoE module at each attention-FFN block, and merging other layers, such as self-attention layers, by simply averaging their weights~\citep{sukhbaatar2024branch}. An additional gate network is introduced for each MoE module to perform token-level routing. However, this approach requires extensive fine-tuning of the entire network parameters and the gating network, as there is a significant gap between MoE experts and pre-trained LLMs. Notably, each expert in an MoE tends to become a generalist across all domains due to the load balancing loss, which encourages an even distribution of the workload among experts~\citep{fedus2022switch,jiang2024mixtral}. This contrasts with pre-trained expert models, which are typically specialized for specific domains. Compared with MoE, our model-level routing has a significantly lower training cost, as we only need to train a router while keeping the expert models frozen.

\noindent\textbf{Model size reduction analysis.} 
Let $\Psi$ be the model size of an FP16 pre-trained model. When storing $M$ FP16 models, the total model size is $M \Psi$. In contrast, using our method, we store a single base model, a model-level router and $M$ compressed delta models. The total model size is $\Psi + M \Tilde{\Psi} + \Phi$, where $ \Tilde{\Psi}$ represents the size of the compressed delta model and $\Phi$ denotes the storage requirement for the router. Therefore, the compression ratio can be computed by $M \Psi / (\Psi + M \Tilde{\Psi} + \Phi)$. For example, when serving 9 LLaMA-13B models, our method achieves a compression ratio of 3.63$\times$. Empirical studies on compression ratios for varying model numbers are shown in Figures~\ref{fig:model_size_reduction} and~\ref{fig:model_size_reduction_llama13b}.

\section{Experiments}
\label{sec:experiments}
\noindent\textbf{Candidate LLMs.} We apply our ME-Switch to two model families, Mistral-7B~\citep{jiang2023mistral} and LLaMA-2-13B~\citep{touvron2023llama}. For the Mistral family, we include Dolphin-2.2.1-Mistral-7B~\footnote{{https://huggingface.co/cognitivecomputations/dolphin-2.2.1-mistral-7b}} as the instruction expert, Speechless-Code-Mistral-7B~\footnote{https://huggingface.co/uukuguy/speechless-code-mistral-7b-v1.0} as the code expert, and MetaMath-Mistral-7B~\footnote{https://huggingface.co/meta-math/MetaMath-Mistral-7B} as the math expert. For the LLaMA-2-13B family, LLaMA-2-13B-Chat~\citep{touvron2023llama} serves as the instruction expert, MetaMath-13B~\footnote{https://huggingface.co/meta-math/MetaMath-13B-V1.0} as the math expert, and LLaMA2-Chinese-13B-Chat~\footnote{https://huggingface.co/FlagAlpha/Llama2-Chinese-13b-Chat} as the Chinese expert. The above models are fine-tuned based on pre-trained backbones. We use Qwen1.5-1.8B-Chat~\citep{qwen} as the model-level router.

\noindent\textbf{Training and testing datasets.}  We collect a diverse set of instruction samples from various open-source datasets, including Alpaca~\citep{alpaca} for the instruction domain, MetaMathQA~\citep{yu2023metamath} for the mathematics domain, Code-74k-ShareGPT~\footnote{https://huggingface.co/datasets/ajibawa-2023/Code-74k-ShareGPT} for the code domain, and Chinese Alpaca~\footnote{https://huggingface.co/datasets/hfl/alpaca\_zh\_51k} for the Chinese domain. To measure the performance of the resulting LLMs, we report accuracy on several benchmarks across different domains: MMLU~\citep{hendrycks2021measuring} for the instruction, GSM8K~\citep{cobbe2021training} and MATH~\citep{hendrycks2021measuringmath} for the mathematics, HumanEval~\citep{chen2021evaluating} and MBPP~\citep{austin2021program} for the code, and C-Eval~\citep{huang2024c} and C-MMLU~\citep{li2023cmmlu} for the Chinese. We use the WizardCoder toolbox to evaluate on HumanEval and MBPP, and the OpenCompass toolbox~\citep{2023opencompass} to evaluate on other datasets. \revise{For MMLU, we report accuracy based on 5-shot in-context learning. To determine the answer for each question, we assess the perplexity of various response options and select the one with the lowest perplexity. For GSM8K and MATH, we adopt a 4-shot Chain of Thought (CoT) methodology to obtain the final answer following~\citep{wei2022chain}. For the HumanEval and MBPP datasets, we employ a 0-shot configuration and generate answers using greedy decoding. We assess the functional correctness using the pass@1 metric following~\citep{liu2024your}.}

\noindent\textbf{Implementation details.} For salient-aware delta compression, we construct a calibration set from each domain-specific dataset and use these sets to compress the delta weights for each respective domain. Each calibration set consists of 1600 randomly sampled sequences, each with a length of 128 tokens. 
The bitwidth $b$ and the number of FP16 input channels $k$ are set to 2 and 8, respectively. We use the AdamW optimizer~\citep{loshchilov2018decoupled} with a learning rate of $10^{-5}$ and a mini-batch size of 4 for training over 1 epoch. Delta weights compression experiments for the Mistral family are conducted on two NVIDIA A100 80G GPUs, while for the LLaMA-2-13B model, we use four NVIDIA A100 80G GPUs. 

For the model-level router training, we construct the training data using samples from various domains, as mentioned in Section~\ref{sec:model_level_routing}. To balance the dataset, we extract an equal number of samples from each domain-specific dataset. This constructed dataset is used to fine-tune model-level router through supervised fine-tuning for 4 epochs on a machine with 8 $\times$ A100 GPUs. We use the AdamW optimizer with $\beta_1=0.9$ and $\beta_2=0.95$, setting the learning rate to $3 \times 10^{-4}$ and applying a linear learning rate warmup. The weight decay is set to 0.01. We set the per-device mini-batch size to 8 and use gradient accumulation steps of 2. 

\begin{table*}[!t]
\renewcommand{\arraystretch}{1.3}
\caption{Main results for Mistral-7B and LLaMA-13B families.}
\vspace{0.1in}
\centering
\scalebox{0.66}
{
\begin{tabular}{cccccc|ccc|cccccc}
\toprule
\multirow{2}{*}{Model} & \multicolumn{5}{c}{MMLU (\%) $\uparrow$} & \multicolumn{3}{c}{Mathematical Reasoning (\%) $\uparrow$} & \multicolumn{3}{c}{Code Generation (\%) $\uparrow$}  \\
\cmidrule(l){2-6} \cmidrule(l){7-9} \cmidrule(l){10-12} 
& STEM & Hums. & Social & Other & Avg. & GSM8K & Math & Avg. & HumanEval & MBPP & Avg. \\
\midrule
Dolphin-2.2.1-Mistral-7B & 52.05 & 68.83 & 73.42 & 65.43 & {63.43} & 63.68 & 12.80 & 38.24 & 42.70 & 54.90 & 48.80 \\
MetaMath-Mistral-7B & 50.45 & 66.82 & 71.63 & 64.60 & 61.87 & 73.92 & 20.62 & \textbf{47.27} & 0.00 & 21.60 & 10.80 \\
Speechless-Code-Mistral-7B & 51.82 & 68.35 & 73.74 & 65.69 & 63.36 & 61.18 & 13.52 & 37.35 & 51.20 & 60.40 & {55.80} \\
\midrule
ME-Switch w/o Routing & 53.17 & 69.09 & 73.88 &	65.40 & \textbf{63.95} & 73.62 & 20.48 & {47.05} & 51.80 & 60.70 & \textbf{56.25} \\
ME-Switch & 51.49 & 68.37 & 73.60 &	66.08 & 63.32 & 73.39 & 20.30 & 46.85 & 51.80 & 60.70 & \textbf{56.25} \\
\midrule

\multirow{2}{*}{Model} & \multicolumn{5}{c}{MMLU (\%) $\uparrow$} & \multicolumn{3}{c}{Mathematical Reasoning (\%) $\uparrow$} & \multicolumn{3}{c}{Chinese (\%) $\uparrow$}  \\
\cmidrule(l){2-6} \cmidrule(l){7-9} \cmidrule(l){10-12} 
& STEM & Hums. & Social & Other & Avg. & GSM8K & Math & Avg. & C-Eval & C-MMLU & Avg. \\
\midrule
LLaMA-2-13B-Chat & 44.26 & 59.79 & 63.20 & 56.57 & 54.60 & 43.75 & 5.20 & 24.48 & 36.13 & 38.71 & 37.42 \\
MetaMath-13B & 37.81 & 52.77 & 56.00 & 50.05 & 47.84 & 69.14 & 8.48 & {38.81} & 33.62 & 32.70 & 33.16 \\
LLaMA2-Chinese-13B-Chat &45.24 & 60.01 & 62.47 & 55.92 & {54.67}& 38.89 & 4.54 & 21.72 & 40.28	& 39.16 & {39.72} \\
\midrule
ME-Switch w/o Routing & 44.57 & 60.87 & 64.00 &	58.04 & \textbf{55.45} & 70.05 & 13.20 & \textbf{41.63} & 40.13 & 39.91 & \textbf{40.02} \\
ME-Switch & 44.51 & 60.87 & 64.00 &	58.04 & {55.43} & 69.90 & 13.14 & 41.52 & 40.13 & 39.84 & 39.99 \\
\bottomrule
\end{tabular}
}
\label{table:results_mistral_llama}
\vspace{-0.15in}
\end{table*}

\vspace{-0.5em}
\subsection{Main Results}
\vspace{-0.5em}
To evaluate the efficacy of our proposed model, we apply ME-Switch to the Mistral-7B and LLaMA-13B model families. \revise{For results in terms of LLaMA-3-8B model family, please see Section~\ref{sec:llama3_8b_family} of the appendix. We include ME-Switch without model-level routing for comparisons.}
The experimental results, detailed in Table \ref{table:results_mistral_llama}, demonstrate that \revise{ME-Switch without routing}, even with extremely compressed delta weights, achieves performance comparable to that of the respective unquantized expert models across various downstream tasks. For the Mistral-7B family, on MMLU, \revise{ME-Switch without routing} 
lags behind the math expert by just 0.22\% in mathematical reasoning tasks. 
Notably, \revise{ME-Switch without routing} consistently outperforms the code expert in code generation tasks. 
\revise{The performance improvements over uncompressed expert models are primarily attributed to additional training through efficient distillation, which improves the models’ task-specific performance by optimizing the quantization step size.}
\revise{With model-level routing, ME-Switch nearly matches the performance of the ones without it, demonstrating its precise routing for user queries.} 

\begin{small}
\begin{table}[!t]
    \begin{minipage}[t]{0.6\linewidth}
        \centering
        \renewcommand{\arraystretch}{1.3}
        \caption{\revise{Performance comparisons between fixed-precision quantization and mixed-precision quantization for MetaMath-Mistral-7B and Speechless-Code-Mistral-7B.}
        }
        \vspace{0.10in}
        \centering
        \scalebox{0.56}
        {
        \begin{tabular}{ccccccccccccccc}
        \toprule
        \multirow{2}{*}{Method} & \multirow{2}{*}{Model Size (GB)} & \multicolumn{3}{c}{Mathematical Reasoning (\%) $\uparrow$} & \multicolumn{3}{c}{Code Generation (\%) $\uparrow$} \\
        \cmidrule(lr){3-5} \cmidrule(lr){6-8} 
        & &  GSM8K & Math & Avg. & HumanEval & MBPP & Avg. \\
        \midrule
        Full-precision  & 13.48 & 73.92 & 20.62 & 47.27 & 51.20 & 60.40 & 55.80 \\
        \cdashline{1-9}
        Fixed-precision & 2.11 & 73.31 & 20.44 & 46.88 & 47.00 & 59.10 & 53.05 \\
        Mixed-precision & 2.13 & {73.62} & {20.48} & \textbf{47.05} & {51.80} & {60.70} & \textbf{56.25}  \\
        \bottomrule
        \end{tabular}
        }
        \label{table:effect_salient_delta_compression}
    \end{minipage}\hfill	%
    \begin{minipage}[t]{0.35\linewidth}
        \centering
        \renewcommand{\arraystretch}{1.3}
        \caption{\revise{Effect of different FP16 input channel numbers $k$ for Speechless-Code-Mistral-7B.}}
        \vspace{0.02in}
        \centering
        \scalebox{0.56}
        {
        \begin{tabular}{ccccccccccccccc}
        \toprule
        Model & Model Size (GB) & HumanEval & MBPP & Avg.\\
        \midrule
        FP & 13.48 & 51.20 & 60.40 & 55.80 \\
        \cdashline{1-5}
        $k=8$ & 2.13 & 51.80 & 60.70 &  56.25 \\
        $k=16$ & 2.15 & 49.40 & 61.90 & 55.65\\
        $k=32$ & 2.19 & 51.20 & 61.20 & 56.20 \\
        $k=64$ & 2.26 & 51.20 & 61.60 & \textbf{56.40} \\
        \bottomrule
        \end{tabular}
        }
        \label{table:effect_different_channel_numbers}
    \end{minipage}
    \vspace{-0.2in}
\end{table}
\end{small}

\vspace{-0.5em}
\subsection{Ablation Studies}
\label{sec:ablation}
\vspace{-0.5em}

\noindent\textbf{\revise{Fixed-precision quantization \vs~Mixed-precision quantization.}}
To validate the effect of \revise{mixed-precision quantization}, we compress the delta weights of MetaMath-Mistral-7B and Speechless-Code-Mistral-7B using both \revise{fixed-precision quantization and our salient-aware mixed-precision quantization. We}
evaluate their performance on mathematical reasoning and code generation tasks, respectively. \revise{The bitwidth $b$ and the number of FP16 input channels $k$ are set to default values as specified in the implementation details.} The results in Table~\ref{table:effect_salient_delta_compression} indicate that despite retaining a minimal number of FP16 channels, the model size of the mixed-precision model (2.13 GB) \revise{is nearly identical} compared to \revise{fixed-precision model} (2.11 GB). 
However, introducing a small number of FP channels significantly improves performance. For example, our compressed Speechless-Code-Mistral-7B, with a model size reduced by 6.33$\times$, even outperforms the full-precision counterpart by 0.45\% in the average accuracy on code generation tasks. This underscores the capability of \revise{salient-aware mixed-precision quantization} to minimize model size while preserving model performance.

\noindent\textbf{\revise{Low-rank adaptation \vs~our salient-aware delta compression.}} \revise{In addition to mixed-precision quantization, we can also employ low-rank adaptation (LoRA) to compress delta weights. Specifically, we decompose the delta weights as $\Delta =\mathbf{U} \mathbf{\Sigma} \mathbf{V}$ and approximate delta weights using low-rank approximation $\tilde{\Delta} = \mathbf{A} \mathbf{B}$ where $\mathbf{A}=\tilde{\mathbf{U}} \sqrt{\tilde{\mathbf{\Sigma}}}$ and $\mathbf{B}=\sqrt{\tilde{\mathbf{\Sigma}}} \tilde{\mathbf{V}}$. Subsequently, $\mathbf{A}$ and $\mathbf{B}$ are refined using our efficient distillation mentioned in Section~\ref{sec:salient_aware_delta_compression}. To compare the effectiveness of LoRA against mixed-precision quantization, we apply both methods to the delta weights of Dolphin-2.2.1-Mistral-7B and MetaMath-Mistral-7B and evaluate performance on instructional and mathematical reasoning tasks. For LoRA, we set the rank to 512. 
The results are detailed in Table~\ref{table:lora_vs_ours}. Our approach with a much smaller model size outperforms LoRA, especially on Math. These results reveal that LoRA cannot accurately approximate delta weights for full fine-tuned models like Dolphin-2.2.1-Mistral-7B and MetaMath-Mistral-7B, given that their delta weights lack low-rank properties, as also shown in Figure~\ref{fig:accumulative_energy} in the appendix.} 

\noindent\textbf{\revise{Effect of different number of FP16 channels.}} \revise{To assess the impact of varying FP16 input channel counts $k$, we compress the delta weights of Speechless-Code-Mistral-7B and evaluate performance on code generation tasks. The bitwidth $b$ is set at 2. From Table~\ref{table:effect_different_channel_numbers}, our method already achieves lossless performance with $k=8$. Increasing $k$ further yields no significant performance improvements, indicating that performance has plateaued. Therefore, we set $k$ to 8 by default.}

\begin{small}
\begin{table}[!t]
        \begin{minipage}[t]{0.67\linewidth}
        \centering
        \renewcommand{\arraystretch}{1.3}
        \caption{\revise{Performance comparisons between LoRA and our salient-aware delta compression for Dolphin-2.2.1-Mistral-7B and MetaMath-Mistral-7B.}}
        \vspace{0.13in}
        \centering
        \scalebox{0.61}
        {
        \begin{tabular}{ccccccccccccccc}
        \toprule
        \multirow{2}{*}{Model} & \multirow{2}{*}{Model Size (GB)} & \multicolumn{5}{c}{MMLU (\%) $\uparrow$} & \multicolumn{3}{c}{Mathematical Reasoning (\%) $\uparrow$} \\
        \cmidrule(l){3-7} \cmidrule(l){8-10}
        & & STEM & Hums. & Social & Other & Avg. & GSM8K & Math & Avg. \\
        \midrule
        LoRA & 2.99 & 52.16 & 68.38 & 73.15 & 65.52 & 63.33 & 73.62 & 17.30 & 45.46 \\
        Ours & \textbf{2.11} & 53.17 & 69.09 & 73.88 & 65.40 & \textbf{63.95} & 73.62 & 20.48 & \textbf{47.05} \\
        \bottomrule
        \end{tabular}
        }
        \label{table:lora_vs_ours}
    \end{minipage}
    \hfill
    \begin{minipage}[t]{0.27\linewidth}
        \centering
        \renewcommand{\arraystretch}{1.3}
        \caption{\revise{Effect of different bitwidths $b$ for Speechless-Code-Mistral-7B.}}
        \vspace{0.05in}
        \centering
        \scalebox{0.61}
        {
        \begin{tabular}{ccccccccccccccc}
        \toprule
        Model & HumanEval & MBPP & Avg.\\
        \midrule
        FP & 51.20 & 60.40 & 55.80 \\
        \cdashline{1-4}
        $b=1$ & 49.40 & 49.90 & 49.65 \\
        $b=2$ & 51.80 & 60.70 & \textbf{56.25} \\
        $b=4$ & 51.80 & 60.20 & 56.00\\
        \bottomrule
        \end{tabular}
        }
        \label{table:effect_different_bitwidths}
    \end{minipage}
    \vspace{-0.05in}
\end{table}
\end{small}

\noindent\textbf{\revise{Effect of different quantization bitwidths.}} \revise{To investigate the impact of varying bitwidths $b$, we compress the delta weights of Speechless-Code-Mistral-7B and evaluate the performance on a code generation task. Table~\ref{table:effect_different_bitwidths} shows that increasing $b$ from 1 to 2 significantly improve performance, achieving lossless results. Given that further increasing $b$ lead to negligible performance differences due to saturation, we set $b$ to 2 as the default.}

\begin{figure}[!t]
	\centering
	\includegraphics[width=0.95\linewidth]{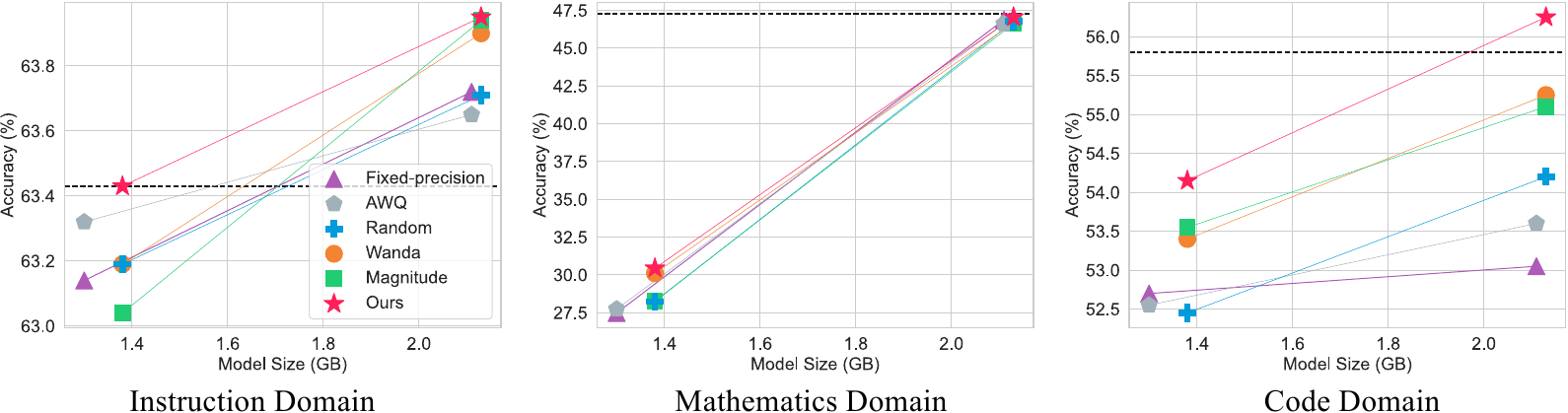}
        \vspace{-0.05in}
	\caption{Average accuracy \vs~delta weights size across different domains. ``Baseline''  refers to the fixed-precision quantization baseline. The dashed line indicates the full-precision counterpart.}
	\label{fig:acc_curve}
    \vspace{-0.2in}
\end{figure}

\noindent\textbf{Performance comparisons with other weight-only quantization methods.}
To demonstrate the promising performance of our salient-aware delta compression, we include the following weight-only quantization methods: \textbf{AWQ}: we use AWQ~\citep{lin2023awq} to rescale input channels of delta weights before quantization to mitigate quantization errors. \textbf{Random}: using our salient-aware delta compression, we randomly select some input channels from the delta weights as important channels. \textbf{Wanda}: leveraging our salient-aware delta compression, we select important input channels of delta weights using the pruning metric from Wanda~\citep{sun2024a}. \textbf{Magnitude}: within our salient-aware delta compression framework, we select sensitive input channels of delta weights based on their weight magnitude, following the method proposed by~\citep{dettmers2022gpt3}. \revise{We also include fixed-precision quantization for comparisons. We applied all methods to compress the delta weights of Dolphin-2.2.1-Mistral-7B, MetaMath-Mistral-7B, and Speechless-Code-Mistral-7B, using bitwidths $b=1$ and $b=2$.} The results are shown in Figure~\ref{fig:acc_curve}. The detailed number of different methods can be found at Section~\ref{sec:different_weight_only_quant} of the appendix.
From the results, we observe that AWQ achieves comparable performance to the 1-bit baseline on code domain, highlighting the limitations of rescaling in extremely low-bitwidth quantization. In contrast, keeping the salient delta input channels performs favourably against the rescaling input channel counterpart. Moreover, our salient channel selection demonstrates superior performance than Random, Wanda, and Magnitude metrics \revise{across various bitwidths and tasks}. For example, \revise{for $b=2$}, our salient-aware delta compression outperforms Wanda by 1.0\% on the average accuracy on code domain, \revise{underscoring the effectiveness and superiority of our approach in selecting the salient delta weights}.

\begin{figure}[!ht]
    \vspace{-0.05in}
    \centering
    \begin{minipage}{.48\textwidth}
        \centering
        \includegraphics[width=0.85\linewidth]{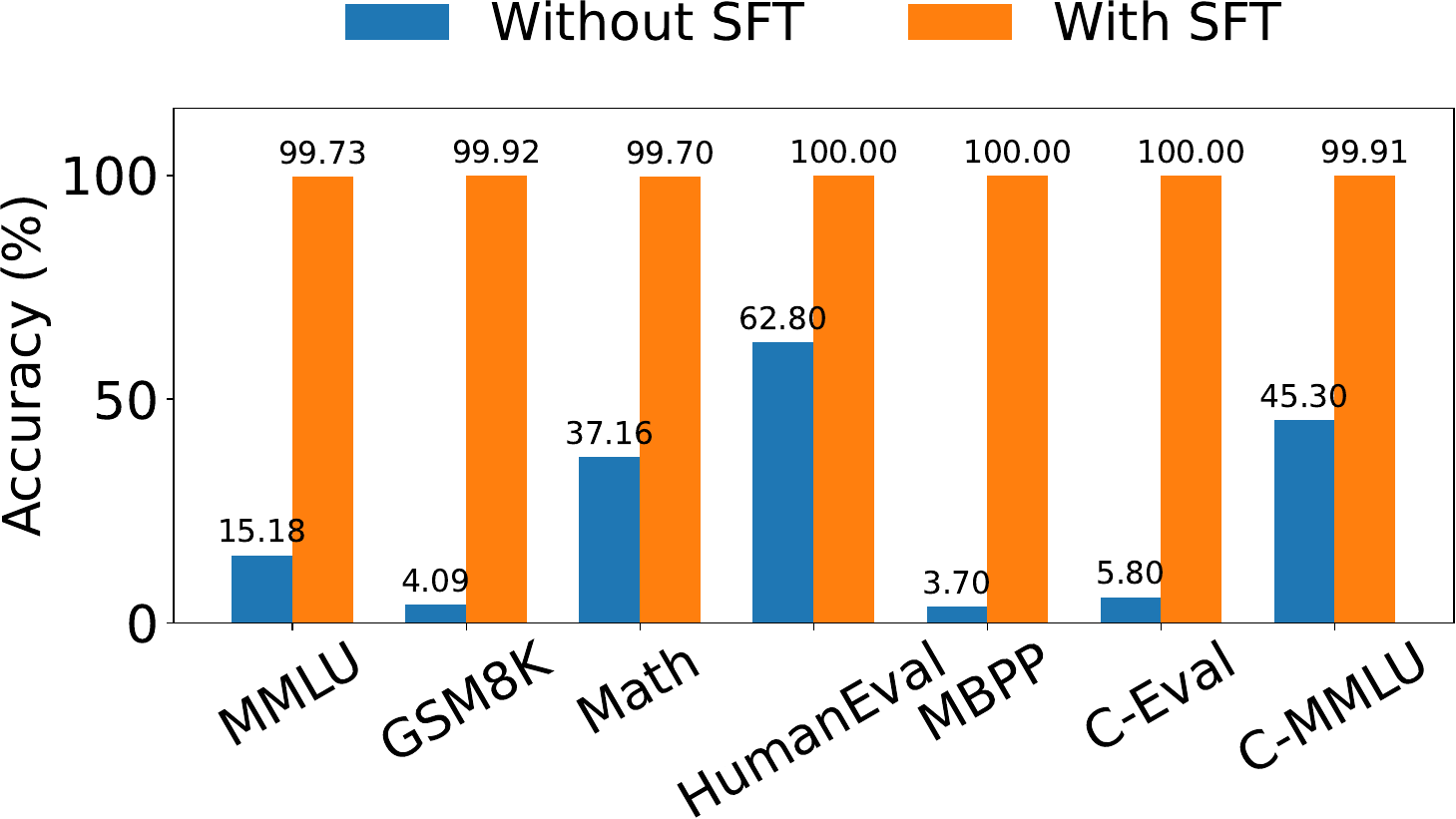}
        \vspace{-0.5em}
        \caption{Effect of supervised fine-tuning (SFT) in model-level routing. We assess the performance of routing by measuring the accuracy on a 4-domain classification task (instruction, mathematics, code, and Chinese).}
        \label{fig:router_performance}
    \end{minipage}%
    \hfill
    \begin{minipage}{0.48\textwidth}
        \centering
        \includegraphics[width=0.85\linewidth]{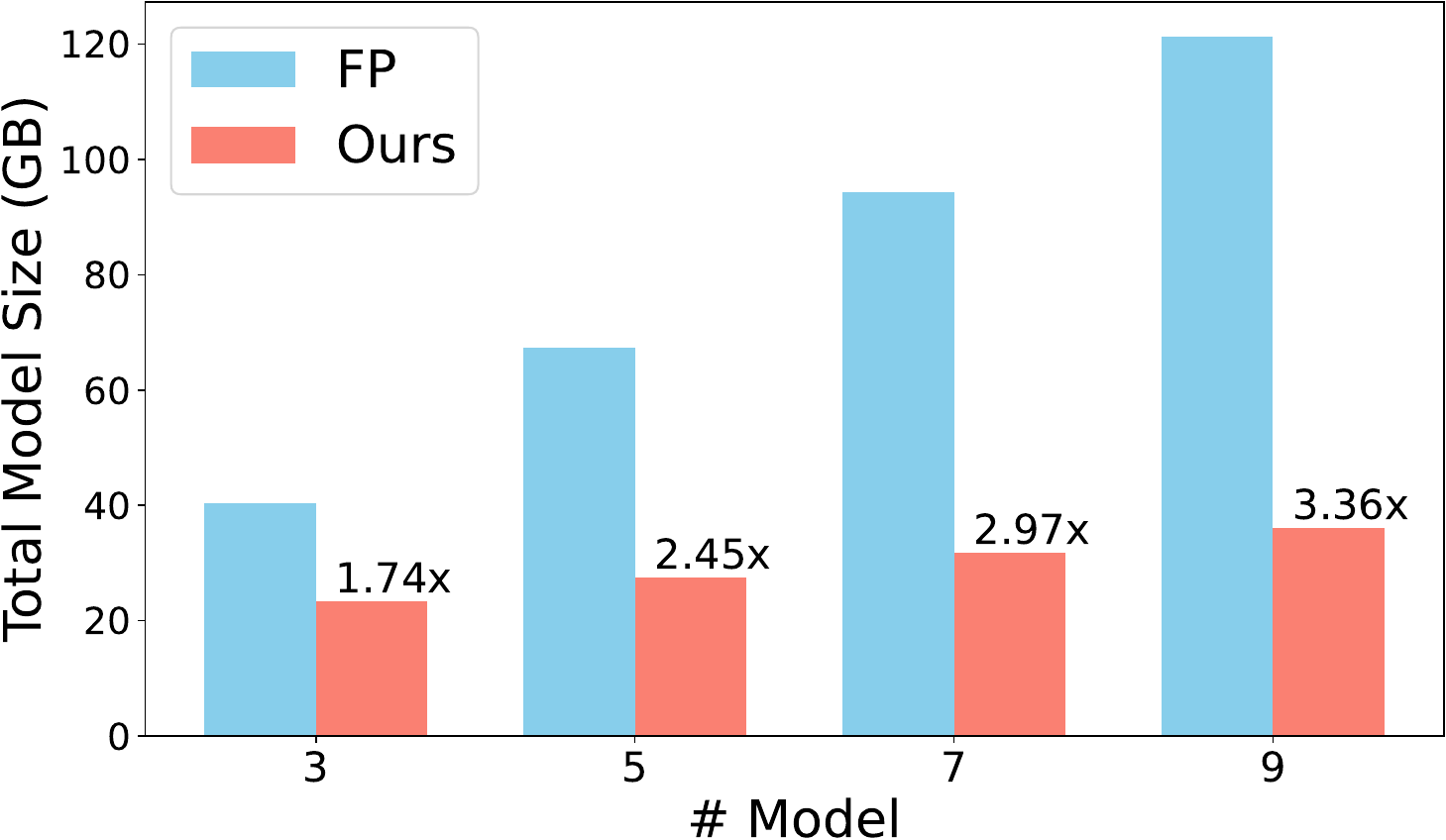}
        \vspace{-0.6em}
        \caption{Model size reduction results in terms of Mistral-7B family. The model sizes for the a single 16-bit floating-point model, a compressed model, and the router are 13.48 GB, 2.13 GB, and 3.42 GB, respectively.
        }
        \label{fig:model_size_reduction}
    \end{minipage}
    \vspace{-0.1in}
\end{figure}

\noindent\textbf{Effect of supervised fine-tuning in model-level routing.}
To investigate the effect of supervised fine-tuning (SFT) on model-level routing, we  evaluate the domain classification performance of the router (\ie, Qwen1.5-1.8B-Chat) across four domains: instruction, mathematics, code, and Chinese. As shown in Figure~\ref{fig:router_performance}, the pre-trained router performs poorly in domain classification without fine-tuning, achieving a Top-1 accuracy of only $5.80\%$ on C-Eval. However, with SFT, the router's performance improves significantly, reaching nearly 100\% accuracy across all domains. This demonstrates that supervised fine-tuning greatly enhances the instruction-following capabilities of the router, thereby improving its routing performance.

\begin{figure}[!ht]
    \vspace{-0.05in}
    \centering
    \begin{minipage}{.5\textwidth}
        \centering
        \includegraphics[width=0.85\linewidth]{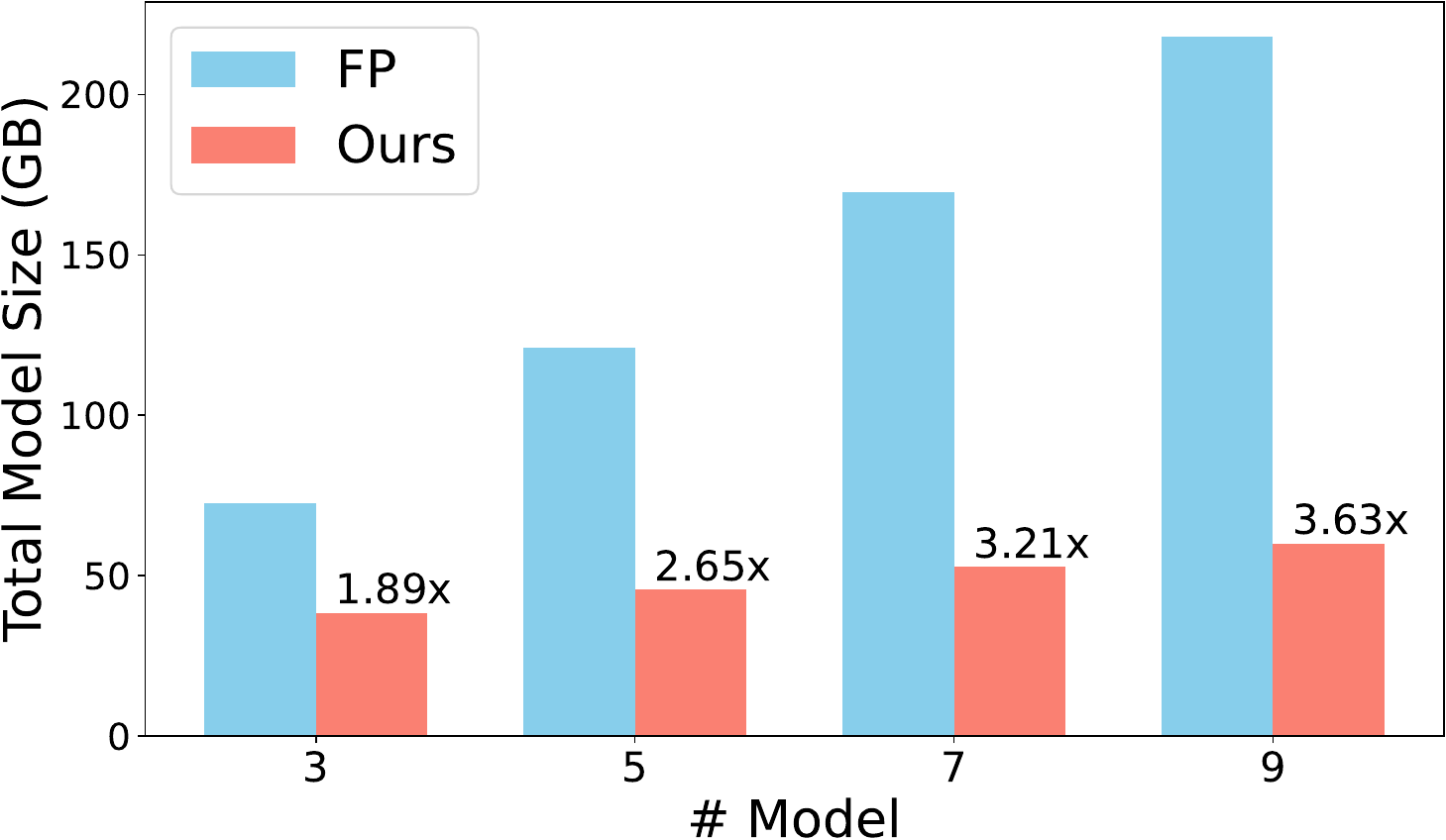}
        \vspace{-0.2em}
        \caption{Model size reduction results in terms of LLaMA-13B family. The model sizes for a single FP16 model, a compressed model, and the router are 24.23 GB, 3.60 GB, and 3.42 GB, respectively.}
        \label{fig:model_size_reduction_llama13b}
    \end{minipage}%
    \hfill
    \begin{minipage}{0.46\textwidth}
        \centering
        \includegraphics[width=0.81\linewidth]{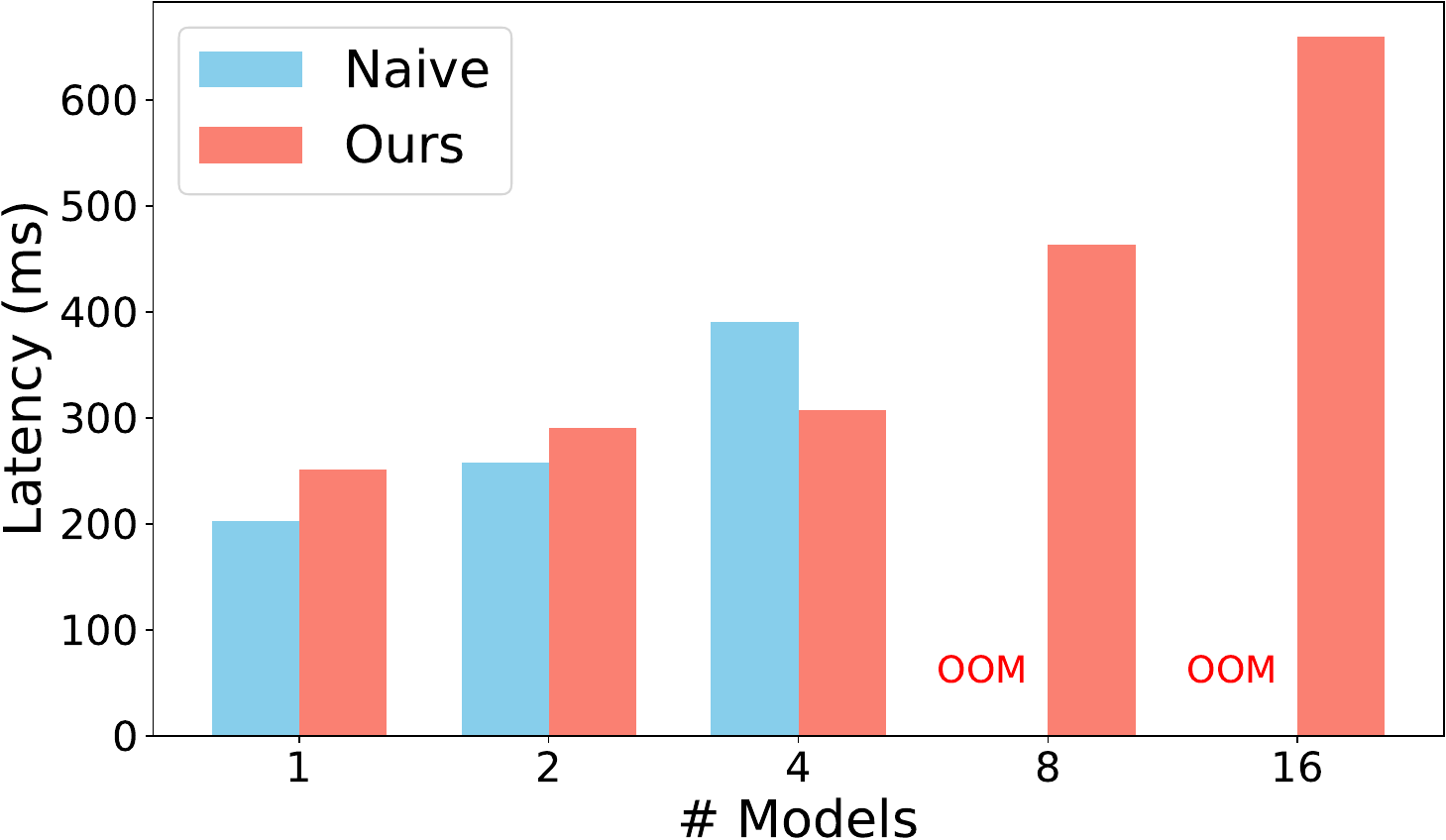}
        \vspace{-0.1em}
        \caption{Decoding latency for Mistral-7B. ``Naive'' denotes the naive inference with $M$ fine-tuned models. ``Ours'' represents batch inference with our method. Out-of-memory scenarios are indicated as ``OOM''.}
        \label{fig:model_latency}
    \end{minipage}
\end{figure}

\noindent\textbf{Model size reduction analysis.} To investigate the model size reduction as discussed in Section~\ref{sec:model_level_routing}, we compare the total storage requirements of full-precision models with those of our compressed models for the Mistral-7B and LLaMA-13B families across varying model counts, as shown in Figures~\ref{fig:model_size_reduction} and~\ref{fig:model_size_reduction_llama13b}. As the number of models increases, the compression ratios improve substantially. For instance, with nine models, our method achieves a 3.63$\times$ reduction compared to full-precision models for Mistral-7B family. These savings become even more pronounced with larger model sizes, reaching up to a 3.63$\times$ reduction for LLaMA-13B family.

\noindent\textbf{Latency analysis.} To assess the latency improvements from delta weights compression, we measured the end-to-end decoding latency of the Mistral-7B model with an input sequence length of 128 on a single NVIDIA A100. \revise{Due to page limitations, additional latency decomposition results are provided in Section~\ref{sec:latency_decomposition} of the appendix.}
Decoding latency is critical, as it typically dominates processing time in LLM operations~\citep{lin2023awq,liu2023deja}. Our efficient Triton kernel, which enables batched matrix multiplication between multiple compressed weight matrices and high-precision input activations, is compared against the conventional approach of individually processing multiple models. 
Results depicted in Figure~\ref{fig:model_latency} illustrate that while our method may perform slightly slower than the naive approach for a small number of models \revise{due to additional dequantization overhead}, it provides lower latency as the number of models is greater than 4. \jing{More importantly, unlike the naive approach, our method is able to simultaneously serve 16 models on GPUs without running into out-of-memory (OOM) issues, demonstrating better scalability and efficiency in high-load scenarios.}

\section{Conclusion and Future Work}
In this paper, we have introduced ME-Switch, a memory-efficient expert switching framework designed for LLMs. Our method has addressed the critical challenge of balancing model performance with storage efficiency. The core of our ME-Switch lies in a novel mixed-precision quantization method that selectively compresses non-salient delta weights to extremely low-bit precision while preserving salient delta weights. Additionally, we have developed a model-level router that dynamically selects the most suitable LLM for a given query by transforming the model selection problem into a domain classification task. Extensive experiments on Mistral-7B and LLaMA-13B families have demonstrated that ME-Switch achieves performance comparable to unquantized expert models across various tasks while significantly reducing model size. In terms of limitations, our ME-Switch relies on accurate domain classification by the model-level router, which can sometimes misclassify and impact performance. Future work could explore more advanced or hybrid routing strategies, such as Mixture of Experts (MoE), to enhance accuracy. Additionally, quantizing the base model itself could further reduce the overall model size. This approach would require careful consideration of the combined effects of quantizing both the base model and the delta weights to ensure performance is maintained. Furthermore, reducing the bitwidth of the KV Cache could accelerate the decoding speed, offering additional efficiency improvements.

\bibliography{iclr2025_conference}
\bibliographystyle{iclr2025_conference}

\appendix
\newpage

\begin{center}
	{
		\Large{\textbf{Appendix}}
	}
\end{center}

\renewcommand\thesection{\Alph{section}}
\renewcommand\thefigure{\Alph{figure}}
\renewcommand\thetable{\Alph{table}}
\renewcommand{\theequation}{\Alph{equation}}

\renewcommand{\theHfigure}{\Alph{figure}}
\renewcommand{\theHtable}{\Alph{table}} %

\setcounter{section}{0}
\setcounter{equation}{0}
\setcounter{figure}{0}
\setcounter{table}{0}

\section{More Details about Prompt Template for Model-level Routing}
\label{sec:prompt_template}
In this section, we presents the prompt template for our model-level routing, as illustrated in Table~\ref{tab:prompt_template}. When a user query is received, it is embedded into the prompt template to form a structured question. This structured question is then processed by the router to perform domain classification.

\begingroup
\begin{table}[H]
    \centering
    \small
    \caption{
    Prompt template for model-level routing. 
    }
    \begin{tabular}{p{\linewidth}}
        \toprule
        \underline{\textbf{\textsc{Prompt template for model-level routing}}} \\
        \vspace{-2mm}
        Classify the query based on the required expertise. Route the query to the appropriate model for a precise response. Only output the letter corresponding to the best category (A, B, C, …, F).
        
        \vspace{1mm}
        Query: \hl{\{Insert the user’s query here. \}}

        \vspace{1mm}
        Options: A) Instruct - For general guidance, explanations, or broad advice. B) Code - For programming-related queries, like debugging or coding. C) Math - For mathematical inquiries, such as problems or theories. D) Chinese Language Expert - For inquiries related to the Chinese language, including translation, grammar, and usage. … F) \hl{\{Specify additional categories and their descriptions here.\}} 

        \vspace{1mm}
        Response should be only 'A', 'B', ‘C’, … or ‘F”, with no additional text. 
        \\
        \bottomrule
    \end{tabular}
    \label{tab:prompt_template}
\end{table}
\endgroup

\section{More results on LLaMA-3-8B model family} 
\label{sec:llama3_8b_family}
\revise{We also applied our method to the LLaMA-3-8B family using the same experimental setup. We employ LLaMA-3-8B-Instruct as the instruction expert and LLaMA-8B-Chinese-Chat as the Chinese expert. The results, presented in Table~\ref{table:results_llama3_8b}, demonstrate that without model-level routing, our ME-Switch achieves nearly lossless performance compared to the unquantized expert models across various downstream tasks, underscoring the superior performance of our salient-aware delta compression. When model-level routing is integrated, performance remains nearly unchanged, highlighting its precise handling of user queries.
}

\begin{table}[!htbp]
    \centering
    \renewcommand{\arraystretch}{1.3}
    \caption{Main results for LLaMA-3-8B family. ``FP Baseline'' refers to the performance metrics of experts without compression.
    }
    \vspace{0.1in}
    \scalebox{0.8}
    {
    \begin{tabular}{cccccc|ccccccccc}
    \toprule
    \multirow{2}{*}{Method} & \multicolumn{5}{c}{MMLU (\%) $\uparrow$} & \multicolumn{3}{c}{Chinese (\%) $\uparrow$} \\
    \cmidrule(l){2-6} \cmidrule(l){7-9} \cmidrule(l){10-12} 
    & STEM & Hums. & Social & Other & Avg. & C-Eval & C-MMLU & Avg. \\
    \midrule 
    FP Baseline & 57.30 & 71.64 & 77.83 & 71.13 & \textbf{68.05} & 51.99 & 52.25 & 52.12 \\
    \cdashline{1-9}
    ME-Switch w/o Routing & 57.12 & 70.89 & 78.60 &	70.92 & 67.93 & 52.67 & 52.70 & \textbf{52.69} \\
    ME-Switch & 57.16 & 70.44 & 78.53 & 71.25 & 67.90 & 52.67 & 52.70 & \textbf{52.69} \\
    \bottomrule
    \end{tabular}
    }
    \label{table:results_llama3_8b}
\end{table}

\section{More performance comparisons with different weight-only quantization methods}
\label{sec:different_weight_only_quant}
\revise{We present the detailed results of Figure~\ref{fig:acc_curve} in Table~\ref{table:comparisons_weight_only}. A comprehensive analysis is available in Section~\ref{sec:ablation}.}

\begin{figure}[!t]
  \begin{center}
    \includegraphics[width=0.98\textwidth]{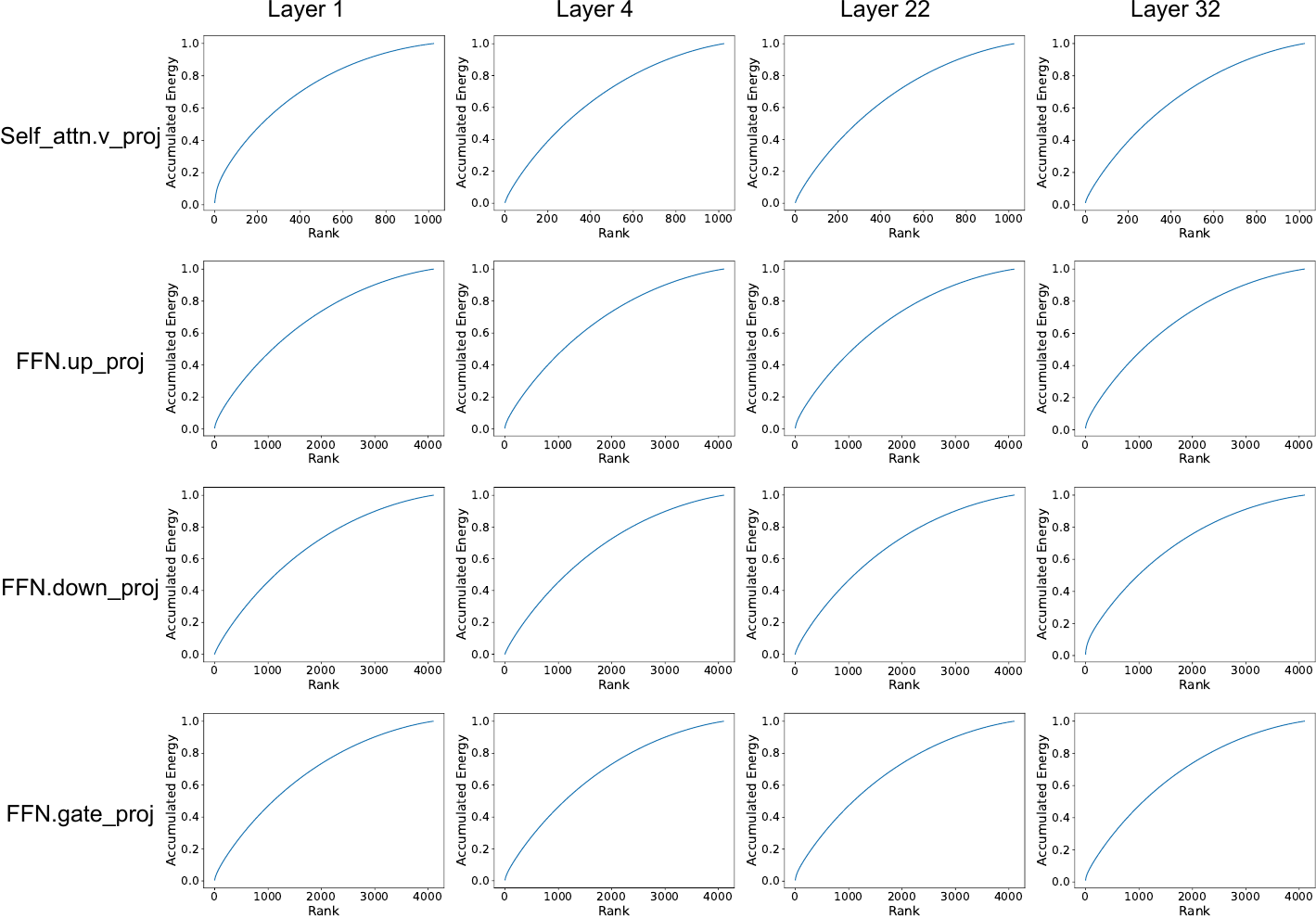}
  \end{center}
  \vspace{-0.15in}
  \caption{\revise{An illustration showing the cumulative energy of delta weights for MetaMath-Mistral-7B model derived through Singular Value Decomposition (SVD).}
  }
  \label{fig:accumulative_energy}
\vspace{-0.1in}
\end{figure}

\begin{table}[!htbp]
    \centering
    \renewcommand{\arraystretch}{1.3}
    \caption{Performance comparisons with different weight-only quantization methods.}
    \vspace{0.1in}
    \scalebox{0.8}
    {
    \begin{tabular}{ccccccccccccccc}
    \toprule
    Domain & Dataset & FP & 1-bit & AWQ & Random & Wanda & Magnitude & Ours \\
    \midrule
    \multirow{1}{*}{\tabincell{c}{Instruct (\%) $\uparrow$}} & MMLU & 63.43 & 63.14 & 63.32 & 63.19 & 63.19 & 63.04 & \textbf{63.43} \\ 
    \midrule
    \multirow{3}{*}{\tabincell{c}{Math (\%) $\uparrow$}} & GSM8K & 73.92 & 53.45 & 53.75 & 54.66 & 58.07 & 54.89 & 59.14 \\
    & Math & 20.62 & 1.50 & 1.72 & 1.82 & 2.14 & 1.64 & 1.74 \\
    & Avg. & 47.27 & 27.48 & 27.74 & 28.24 & 30.11 & 28.27 & \textbf{30.44} \\
    \midrule
    \multirow{3}{*}{\tabincell{c}{Code (\%) $\uparrow$}} & HumanEval & 51.20 & 47.00 & 47.00 & 46.30 & 48.20 & 48.20 & 47.60 \\
    & MBPP & 60.40 & 58.40 & 58.10 & 58.60 & 58.60 & 58.90 & 60.70 \\
    & Avg. & 55.80 & 52.70 & 52.55 & 52.45 & 53.40 & 53.55 & \textbf{54.15} \\
    \midrule
    Domain & Dataset & FP & 2-bit & AWQ & Random & Wanda & Magnitude & Ours \\
    \midrule
    \multirow{1}{*}{\tabincell{c}{Instruct (\%) $\uparrow$}} & MMLU & 63.43 & 63.72 & 63.65 & 63.71 & 63.90 & 63.94 & \textbf{63.95} \\
    \midrule
    \multirow{3}{*}{\tabincell{c}{Math (\%) $\uparrow$}} & GSM8K & 73.92 & 73.31 & 73.24 & 73.01 & 72.71& 73.16 & 73.62 \\
    & Math & 20.62 & 20.44 & 19.98 & 20.52 & 20.64 & 20.08 & 20.48 \\
    &  Avg. & 47.27 & 46.88  & 46.61 & 46.77 & 46.68 & 46.62 & \textbf{47.05} \\
    \midrule
    \multirow{3}{*}{\tabincell{c}{Code (\%) $\uparrow$}} & HumanEval & 51.20 & 47.00 & 47.00 & 48.80 & 50.60 & 50.00 & 51.80 \\
    & MBPP & 60.40 & 59.10 & 60.20 & 59.60 & 59.90 & 60.20 & 60.70 \\
    & Avg. & 55.80 & 53.05 & 53.60 & 54.20 & 55.25 & 55.10 & \textbf{56.25} \\
    \bottomrule
    \end{tabular}
    }
    \label{table:comparisons_weight_only}
\end{table}

\section{More analysis on the rank of delta weights}
\label{sec:analysis_rank_delta_weights}
\revise{
We show the cumulative energy of delta weights for MetaMath-Mistral-7B in Figure~\ref{fig:accumulative_energy}, using squared singular values to measure the ``energy'' of the projection matrix. The results show that all projection layers consistently exhibit a similar trend and possess a relatively high rank. Therefore, due to the absence of low-rank properties, LoRA cannot accurately approximate the delta weights of full fine-tuned models.}

\section{BERT \vs~small LLM for model-level routing}
\revise{In addition to smaller LLMs, we can also use BERT for domain classification given user queries. Specifically, we employ DistillBERT~\citep{sanh2019distilbert} as the backbone and fine-tune it on our collected dataset (described in Section~\ref{sec:model_level_routing}) with just the queries and corresponding domain labels. We show the router accuracy in Table~\ref{tab:router_performance_comparisons} and the latency comparisons in Table~\ref{tab:latency_comparisons}. From the results, DistillBERT with a faster response time performs well across most datasets, except for MMLU where its limited capacity struggles with complex data. In contrast, our method consistently achieves good performance across all datasets, demonstrating its effectiveness even in complex scenarios. Moreover, our router’s inference latency is just 17.60ms for sequence lengths of 128, which is less than 7\% of the inference time for expert models. Therefore, we continue to leverage Qwen1.5-1.8B-Chat for its proven effectiveness in challenging scenarios.}

\begin{table}[!tbp]
    \centering
    \renewcommand{\arraystretch}{1.3}
    \caption{Router performance comparisons.}
    \vspace{0.1in}
    \centering
    \scalebox{0.85}
    {
    \begin{tabular}{ccccccccccccccc}
    \toprule
    Model & MMLU & GSM8K & Math & HumanEval & MBPP & C-Eval & C-MMLU \\
    \midrule
    DistillBERT & 73.29 & \textbf{100.00} & \textbf{99.80} & 96.34 & \textbf{100.00} & 99.97 & \textbf{100.00} \\
    Qwen1.5-1.8B-Chat & \textbf{99.73} & 99.92 & 99.70 & \textbf{100.00} & \textbf{100.00} & \textbf{100.00} & 99.91 \\
    \bottomrule
    \end{tabular}
    }
    \label{tab:router_performance_comparisons}
\end{table}

\begin{table}[!tbp]
    \centering
    \renewcommand{\arraystretch}{1.3}
    \caption{Router latency (ms) comparisons.}
    \vspace{0.1in}
    \scalebox{0.9}
    {
    \begin{tabular}{ccccccccccccccc}
    \toprule
    Sequence Length & 128 & 256 & 512 \\
    \midrule
    DistillBERT & \textbf{3.70} & \textbf{3.80} & \textbf{4.90} \\ 
    Qwen1.5-1.8B-Chat & 17.60 & 18.30 & 19.10 \\
    \bottomrule 
    \end{tabular}
    }
    \label{tab:latency_comparisons}
\end{table}

\section{More results on latency decomposition}
\label{sec:latency_decomposition}
\begin{figure}[!h]
	\centering
	\includegraphics[width=0.5\linewidth]{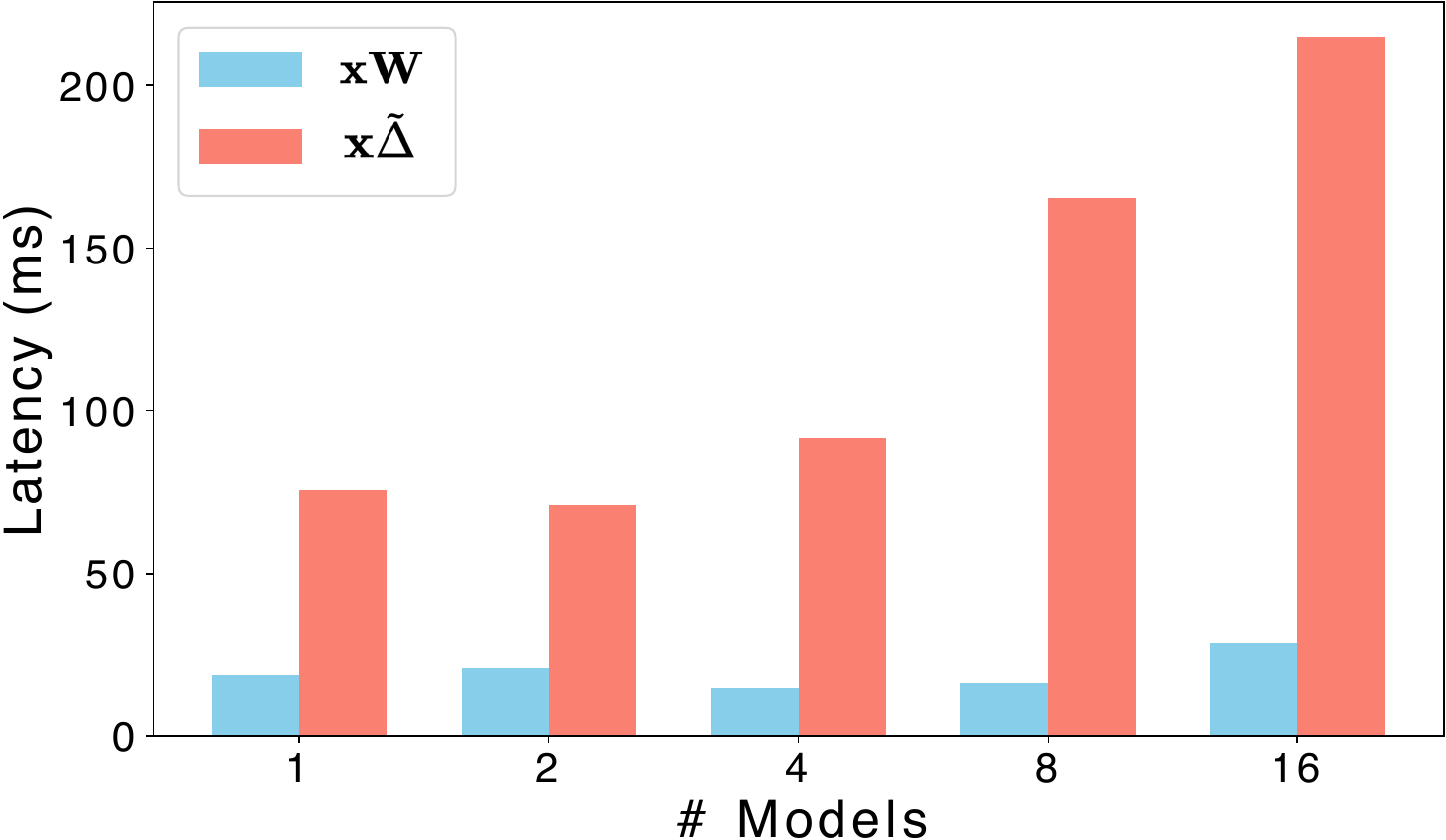}
        \vspace{-0.05in}
	\caption{Latency decomposition of our method for Mistral-7B on a single NVIDIA A100 80G GPU.}
	\label{fig:latency_decomposition}
    \vspace{-0.1in}
\end{figure}

\revise{We provide a detailed breakdown of decoding times for Mistral-7B model in Figure~\ref{fig:latency_decomposition}, using an input sequence length of 128 on a single NVIDIA A100 GPU. Since loading delta weights, dequantization, and matrix multiplication are integrated into our efficient Triton kernel, it is challenging to isolate and measure the latency of each individual component separately. Therefore, we measure the inference time for the operations $\mathbf{x}\mathbf{W}$ and $\mathbf{x}\tilde{\Delta}$. Notably, as the number of models increases, the increased latency attributed to $\mathbf{x}\tilde{\Delta}$ exceeds that of $\mathbf{x}\mathbf{W}$, primarily due to the augmented loading time of the more compressed delta weights.}

\end{document}